\pdfoutput=1
\documentclass[11pt]{article}

\usepackage[letterpaper,margin=1in,headheight=14pt,headsep=18pt]{geometry}
\usepackage[T1]{fontenc}
\usepackage[utf8]{inputenc}
\usepackage{mathptmx}                 
\usepackage[scaled=0.92]{helvet}      
\usepackage{amsmath}
\usepackage[final]{microtype}
\usepackage{xcolor}
\definecolor{ecblue}{HTML}{204ECF}    
\definecolor{ecnavy}{HTML}{0F256E}    
\definecolor{ecgreen}{HTML}{00CC83}   
\definecolor{ecchar}{HTML}{111827}    
\definecolor{ecslate}{HTML}{667085}   
\definecolor{eccool}{HTML}{D9DFEA}    
\definecolor{ecpale}{HTML}{EAF0FF}    
\definecolor{eclight}{HTML}{F4F6FA}   

\usepackage{graphicx}
\usepackage{booktabs,tabularx,array,makecell,threeparttable}
\usepackage{enumitem}
\usepackage{titlesec}
\usepackage{fancyhdr}
\usepackage[font=small,labelfont={bf,sf,color=ecnavy},labelsep=period,
            justification=justified,singlelinecheck=false]{caption}
\usepackage[section]{placeins}
\usepackage{fvextra}
\usepackage[most]{tcolorbox}
\usepackage[authoryear,square]{natbib}
\setcitestyle{aysep={,},yysep={,},citesep={;}}
\usepackage[hyphens]{xurl}
\usepackage[colorlinks=true,linkcolor=ecblue,citecolor=ecblue,urlcolor=ecblue,
            pdftitle={ExpertCrowd-SWE: Measuring Frontier Coding Agents on Expert-Authored Engineering Tasks in Unfamiliar Codebases},
            pdfauthor={Josip Vrdoljak, Akash Singh, Bas Steunebrink, Alex Yarotsky, John Lorenzo, Achille Nicoletti, Sam Pigott}]{hyperref}

\titleformat{\section}{\sffamily\Large\bfseries\color{ecnavy}}{\thesection}{0.75em}{}
\titleformat{\subsection}{\sffamily\large\bfseries\color{ecnavy}}{\thesubsection}{0.75em}{}
\titlespacing*{\section}{0pt}{20pt plus 4pt minus 2pt}{8pt plus 2pt}
\titlespacing*{\subsection}{0pt}{14pt plus 3pt minus 2pt}{5pt plus 2pt}
\newcommand{\runin}[1]{\textbf{#1}\hspace{0.35em}}   
\newcommand{\secref}[1]{Sec.~\ref{#1}}
\newcommand{\aff}[1]{\textsuperscript{\normalfont #1}}          

\setlist{itemsep=3pt,topsep=4pt,parsep=0pt}
\setlist[itemize]{leftmargin=1.4em,label={\color{ecblue}\small\textbullet}}

\renewcommand{\arraystretch}{1.18}
\newcolumntype{L}{>{\raggedright\arraybackslash}X}
\newcommand{\colhead}[1]{\textbf{\sffamily #1}}
\newcommand{\tool}[1]{\texttt{\footnotesize #1}}

\fancypagestyle{firstpage}{\fancyhf{}%
  \fancyfoot[C]{\sffamily\footnotesize\color{ecslate}\thepage}}

\begin{document}

\thispagestyle{firstpage}
\noindent
\begin{minipage}[b]{0.5\textwidth}
  \includegraphics[height=0.78cm]{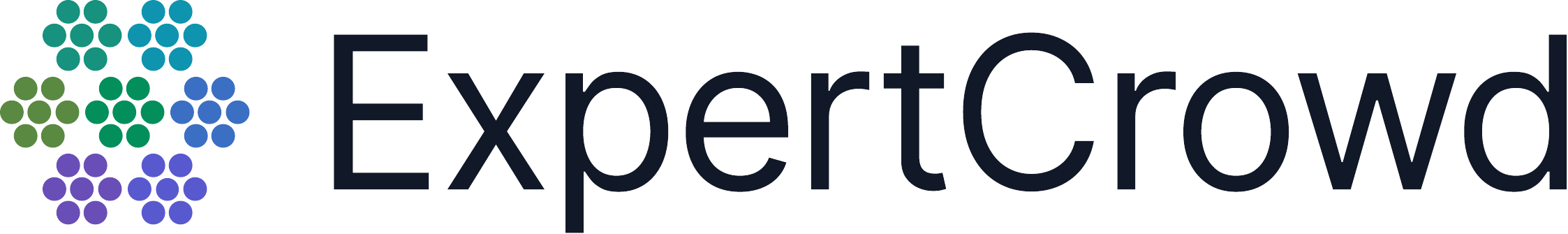}
\end{minipage}%
\hfill
\begin{minipage}[b]{0.45\textwidth}
  \raggedleft\sffamily\footnotesize\color{ecslate}
  Technical report\\November 2026
\end{minipage}

\vspace{7pt}
\noindent{\color{eccool}\rule{\textwidth}{0.6pt}}

\vspace{26pt}
\begin{center}
  {\sffamily\bfseries\fontsize{19}{24}\selectfont\color{ecnavy}
    ExpertCrowd-SWE: Measuring Frontier Coding Agents\\
    on Expert-Authored Engineering Tasks\\
    in Unfamiliar Codebases\par}

  \vspace{18pt}
  {\large\bfseries
    Josip Vrdoljak\aff{1} \qquad Akash Singh\aff{1} \qquad Bas Steunebrink\aff{1} \qquad Alex Yarotsky\aff{1}\\[4pt]
    John Lorenzo\aff{1} \qquad Achille Nicoletti\aff{1} \qquad Sam Pigott\aff{1}\par}

  \vspace{9pt}
  {\normalsize
    \aff{1}Toptal ExpertCrowd\par}
\end{center}

\vspace{14pt}
\begin{center}
\begin{minipage}{0.9\textwidth}
  {\sffamily\bfseries\color{ecnavy}Abstract}\par\vspace{4pt}
  \small
  ExpertCrowd-SWE (EC-SWE) is a benchmark of 100 expert-authored, multi-file software engineering tasks for evaluating autonomous coding agents in large, unfamiliar codebases. Each task is a change request written by a practising engineer against a pinned commit of a real repository, paired with a sealed reference solution and a task-specific grader whose weighted, mostly executable criteria award partial credit; must-pass blockers and pass/fail gates encode what a maintainer would refuse to merge. The tasks span 87 open-source projects and 11 languages. The median ticket is 148 words, while the median reference solution changes 471 lines. Every task image is calibrated before release so that the reference solution scores 1.0 and both the untouched repository and an inert change score 0; nearly half of the tasks also ship alternative, lazy and adversarial control patches that the grader must classify correctly. The final 100 tasks were selected for difficulty against current frontier models. We evaluate seven frontier models, five runs each (3,495 scored runs), in a single platform harness that gives each model its provider-native tool interface. Claude Opus 5.5 (67.3\%) and GPT-6 Astra (63.7\%) lead without a statistically significant gap, followed by Claude Fable 5.1 (46.5\%), Gemini 3.8 Flash (40.0\%), and a statistically indistinguishable group of Qwen 3.8 Max, DeepSeek V4.1 Flash and Grok 4.6 (32--34\%). No task is fully solved by every model, and on 80 tasks no model earns full credit on all five runs. Scores span 35 points. Depending on the model, agents take a median of 33 to 166 steps per run, with little relation to score. We release 24 tasks with their environments, graders and results, and hold out 76 to limit contamination.

  \vspace{8pt}
  {\sffamily\footnotesize\renewcommand{\arraystretch}{1.25}%
  \begin{tabular}{@{}l@{\hspace{0.6em}}l@{}}
    \textbf{\color{ecnavy}Dataset} & \href{https://huggingface.co/datasets/ExpertCrowd/ec-swe}{huggingface.co/datasets/ExpertCrowd/ec-swe} \\
    \textbf{\color{ecnavy}Leaderboard} & \href{https://www.toptal.com/expertcrowd/benchmarks/expertcrowd-swe-benchmark}{toptal.com/expertcrowd/benchmarks/expertcrowd-swe-benchmark}
  \end{tabular}\par}
\end{minipage}
\end{center}

\vspace{6pt}
\noindent{\color{eccool}\rule{\textwidth}{0.6pt}}

\section{Introduction}\label{sec:intro}

Large language models (LLMs) now act as autonomous software engineers: given a request, they explore an unfamiliar repository, edit across files, run builds and tests, and iterate until the change works. Benchmarks have become the standard way to measure these systems and to decide which of them to trust. They can only play that role if their tasks resemble real engineering work, their grading accepts every correct solution and rejects incorrect ones, and their solutions are not already present in the models' training data.

The dominant recipe, introduced by SWE-bench \citep{jimenez2024swebench}, mines merged fixes from public GitHub repositories and grades each task with the tests that shipped with its fix. The recipe scales, but it has two limitations. First, the fixes, their discussion and their tests are public and plausibly in pretraining corpora, so a high score can reflect recall rather than problem solving \citep{liang2025illusion}. Second, inherited tests were written to confirm one particular patch, not to grade arbitrary new solutions, so they can reject valid alternatives or accept incomplete ones. In 2026 both limitations became impossible to ignore. In February, OpenAI stopped reporting SWE-bench Verified after an audit found material issues in the tests or problem statements of 59.4\% of 138 audited problems, together with evidence that frontier models had memorized gold patches \citep{openai2026a}. In July, a follow-up audit estimated that roughly 30\% of SWE-Bench Pro tasks were broken; OpenAI retracted its earlier recommendation of that benchmark and called for new evaluations built with experienced software developers and human oversight throughout \citep{openai2026b}.

Benchmarks with authored rather than mined tasks address the contamination problem \citep{huang2026deepswe,merrill2026terminalbench,kottamasu2026apexswe}, but the strongest agents now crowd the top of those leaderboards as well. On the DeepSWE v1.1 leaderboard, Claude Opus 5.5, GPT-6 Astra, DeepSeek V4.1 Flash and Gemini 3.8 Flash sit within 0.9 points of one another (72.3\% to 71.4\%) \citep{deepswe2026leaderboard}. A benchmark whose top scores are this close cannot tell configurations apart, however well its tasks are built.

What is missing is a benchmark of changes no model has seen solved: written for the benchmark by engineers who know the codebase, never published as fixes, graded the way a maintainer reviews a pull request, and kept mostly private.

This paper introduces \textbf{ExpertCrowd-SWE (EC-SWE)}, a benchmark of 100 multi-file software engineering tasks authored by practising engineers from Toptal's ExpertCrowd network. The tasks are built and evaluated on HUD \citep{hud2026sdk}, a platform for building and evaluating AI agents: each task ships as a container image with its grader, agents run against it through a hosted runner and inference gateway, and every run is recorded as a full trace (\secref{sec:protocol}). Our contributions are:

\begin{itemize}
  \item \textbf{Expert-authored change requests in codebases the authors know.} Each task is a ticket that an engineer who knows the codebase well writes against a pinned commit of a real repository. Most tasks are new changes on top of recent public commits; a few build on code the expert added or owns. Authors are forbidden to base a task on a bug that has already been fixed in public (\secref{sec:recruit}--\ref{sec:sources}).
  \item \textbf{Rubric grading with partial credit.} Each task has a weighted grader written by its author. Executable criteria are checked through public interfaces, and must-pass \emph{blockers} and pass/fail \emph{gates} enforce conditions a maintainer would insist on, such as leaving the repository's own tests untouched. LLM-judged criteria play a deliberately small role (\secref{sec:grading}).
  \item \textbf{Calibrated, expert-checked graders.} A task image only builds if the reference solution scores exactly 1.0 and both the untouched repository and an inert change score 0. Nearly half of the tasks also ship explicit alternative, lazy and adversarial control patches that the grader must classify correctly, and every task is quality-checked by a senior engineer before release (\secref{sec:qa}).
  \item \textbf{Contamination controls.} Each task's change is unpublished, 76 of the 100 tasks are held out, and every task was screened for overlap with 93 public benchmark datasets (\secref{sec:contam}).
  \item \textbf{A seven-model evaluation with reliability and effort data.} We report 3,495 scored runs of seven frontier models, with task-level and run-to-run uncertainty, paired comparisons, full-credit rates, and per-run step counts (\secref{sec:results}).
\end{itemize}

\runin{Scope.}EC-SWE measures one capability: given a short ticket, can an agent working autonomously in a sandboxed, offline repository produce a change that an expert maintainer's grader accepts? All models run through one platform harness that exposes each provider's native tool interface (\secref{sec:protocol}). The leaderboard therefore compares models under shared task, prompt, sandbox and budget conditions. It is not a ranking of coding products or of overall model quality.

\section{Related Work}\label{sec:related}

\runin{The SWE-bench lineage.}SWE-bench framed repository-level evaluation as resolving GitHub issues, with 2,294 instances mined from merged pull requests in twelve Python projects \citep{jimenez2024swebench}. SWE-bench Verified distilled it to 500 human-screened instances and became the de facto industry leaderboard \citep{openai2024verified}. Later work extended the format to more languages \citep{zan2025multiswebench}, to visual software domains \citep{yang2025multimodal}, and to longer, enterprise-style problems with public, held-out and commercial splits \citep{deng2025swebenchpro}. SWE-smith synthesizes task instances at scale for training and evaluation \citep{yang2025swesmith}. EC-SWE keeps this lineage's framing of repository-level, execution-graded tasks, but it authors its tasks rather than mining them and grades them with author-written rubrics rather than inherited tests.

\runin{Authored, non-mined tasks.}A growing line of work writes tasks directly. Aider's polyglot benchmark curates 225 short exercises \citep{gauthier2024aider}. Terminal-Bench authors terminal tasks across many domains, software engineering among them \citep{merrill2026terminalbench}. APEX-SWE has experts author integration and observability tasks \citep{kottamasu2026apexswe}. FrontierSWE curates a small number of very large, open-ended problems graded with partial credit \citep{chu2026frontierswe}. DeepSWE is the closest neighbour: 113 original, never-merged tasks across 91 repositories, graded by hand-written functional verifiers under a single bash-only harness \citep{huang2026deepswe}. EC-SWE shares DeepSWE's emphasis on original tasks at repository scale. It differs in three ways: grading is rubric-based with partial credit rather than pass/fail; the authors are domain experts working in codebases they know; and the released set is explicitly selected for difficulty against current frontier models.

\runin{Contamination and decontamination.}Direct probing shows that frontier models reproduce SWE-bench gold patches far more often than matched off-benchmark code \citep{liang2025illusion}, and OpenAI's red-teaming found every tested frontier model able to reproduce gold patches or verbatim problem details for some SWE-bench Verified tasks \citep{openai2026a}. Defences fall into three families. \emph{Continuous collection} restricts evaluation to tasks created after a model's training cutoff \citep{zhang2025swebenchlive,badertdinov2025swerebench,jain2025livecodebench}. \emph{Access control} holds out private or commercial splits \citep{deng2025swebenchpro}. \emph{Original authorship} keeps reference solutions out of the public record \citep{huang2026deepswe}. EC-SWE combines original authorship with a held-out split (76 of 100 tasks are private) and an explicit similarity check against 93 public benchmark datasets (\secref{sec:contam}).

\runin{Verification and grading.}Most mined benchmarks grade with the resolving commit's \texttt{FAIL\_TO\_PASS} and \texttt{PASS\_TO\_PASS} tests \citep{jimenez2024swebench,openai2024verified}. Audits find these tests too strict, rejecting valid alternatives, and too loose, accepting incomplete fixes \citep{openai2026a,openai2026b}. DeepSWE replaces them with functional verifiers written from the task specification and audits them with an independent LLM judge \citep{huang2026deepswe}. EC-SWE likewise writes its graders from the specification, but it decomposes each task into weighted criteria. Partial credit separates an agent that solves most of a change from one that solves little, and blockers and gates encode the conditions under which a maintainer would reject a change outright.

\runin{Harnesses and judges.}The scaffold between model and repository varies widely, from structured agent-computer interfaces \citep{yang2024sweagent} and general agent platforms \citep{wang2025openhands} to fixed localize-repair pipelines \citep{xia2024agentless} and minimal bash-only loops \citep{sweagent2025mini}. Because the same model can score differently across scaffolds, harness choice confounds cross-model comparison. DeepSWE removes this confound by giving every model a single bash tool \citep{huang2026deepswe}. EC-SWE instead runs every model through one platform harness that exposes each provider's native tool interface, as discussed in \secref{sec:protocol} and \secref{sec:limitations}. LLM-as-a-judge methods \citep{zheng2023judging} and agent-as-a-judge methods \citep{zhuge2024agentjudge} are increasingly used to grade qualities that tests cannot capture. EC-SWE uses them sparingly, for maintainer-style review criteria in a minority of tasks. Finally, we follow recent calls to report uncertainty from task sampling as well as run-to-run noise \citep{madaan2024variance,merrill2026terminalbench}.

\section{Methodology}\label{sec:method}

EC-SWE is built in five stages. Engineers propose and build tasks in codebases they know (\secref{sec:recruit}--\ref{sec:anatomy}). Each task receives a rubric grader (\secref{sec:grading}), which is calibrated and reviewed (\secref{sec:qa}) and checked for contamination (\secref{sec:contam}). Tasks are then selected for difficulty and evaluated under a fixed protocol (\secref{sec:protocol}--\ref{sec:metrics}).

\subsection{Design principles}\label{sec:principles}

Mined benchmarks (\secref{sec:related}) grade a change by whether its tests pass \citep{jimenez2024swebench,openai2024verified}. Our authoring brief instead asks the question a maintainer asks of a pull request, ``would you merge it?'', which takes in correctness, regressions, scope, test quality and maintainability. Every task must be:

\begin{itemize}
  \item \textbf{Realistic:} a change that a real team would ask for, written the way a senior engineer would assign it, saying what is wanted and never how to do it;
  \item \textbf{Solvable:} the author's own reference solution proves the task can be done;
  \item \textbf{Gradable:} a machine can tell a right answer from a wrong one;
  \item \textbf{Reproducible:} the same environment gives the same result.
\end{itemize}

Difficulty must come from reasoning, never from withheld information. We have used four valid sources of difficulty: cross-module synthesis, real trade-offs (going faster costs something the task says must be kept), unwritten but discoverable conventions of the repository, and chained work in which an early decision shapes later steps. We have also specified four invalid ones: facts that appear nowhere in the repository, a single acceptable implementation that the prompt does not signal, a single file that gives the whole answer away, and graders or reference solutions the agent can read. Authors are encouraged to combine two kinds of work, for example performance $\times$ error handling, concurrency $\times$ logging, migration $\times$ backwards compatibility, or refactoring $\times$ test quality, because a task that exercises only one is usually too easy.

\subsection{Expert recruitment and authoring workflow}\label{sec:recruit}

Task authors are engineers from Toptal's talent network (more than 30,000 vetted and specialized experts). Authors apply to one of four areas: compilers and language tooling; systems, backend and networking; database and storage systems; and SRE, DevOps and infrastructure. Thirty-nine experts contributed to task authoring. After QC, verification and selection, the final 100 tasks come from 17 of them.

Authoring follows a two-stage, draft-first workflow:

\begin{itemize}
  \item \textbf{Draft.} The author submits a short, human-readable draft with three parts: the repository and the exact commit hash it is pinned to (never a branch name), the prompt written as a ticket, and a grading plan naming the checks and which of them must pass. A reviewer reads the draft and either approves it or sends it back. Nothing is built before approval, and ``almost'' counts as not approved. Drafts are rejected for being unreadable, for prescribing an implementation, for lacking a grading plan or relying only on an AI judge, or for proposing a problem the author has not solved.
  \item \textbf{Build.} After approval, the author builds the task on the HUD coding template (environment, reference solution and grader), applies their own reference solution, confirms that it scores 100\% while an unchanged checkout scores 0\%, and runs the task five times against a frontier model (Claude Opus 5, step limit 100). A result of 0\% must be checked, since five zero runs also look like a broken grader. AI assistance is permitted for environment wiring but not for the task idea or the reference solution.
\end{itemize}

Recorded reasons for rejecting submitted tasks include reference solutions that were never tested (4), weak, broken or opinion-only graders (3), tasks or repositories that were too small (3), incomplete submissions (2) and one impossible task.

\subsection{Task sources}\label{sec:sources}

Tasks come from four routes, all governed by the same contamination rules: never base a task on a bug that has already been fixed in public, never use employer or client code in any form, keep the task's own change unpublished, and use proprietary code only when the author owns it and grants the rights.

\begin{itemize}
  \item \textbf{Expert experience on public repositories.} The author proposes a change they have shipped, or wished someone had, in a public project they know well, pinned to a recent upstream commit.
  \item \textbf{Long-standing unresolved issues.} The task is motivated by an open, unresolved upstream issue, but the reference solution is the author's own and does not exist upstream.
  \item \textbf{Forks with expert-added problems.} The author forks a public repository and adds unpublished commits or a substrate patch on top. The task is then defined against that added code.
  \item \textbf{Codebases the expert owns.} The task is built on a codebase that the author wrote and controls.
\end{itemize}

In the released set, every task's base commit could be fetched from the GitHub repository named in its environment. Ninety-eight tasks are new changes on public open-source projects with no added substrate. One task (a PostgreSQL foreign-key migration preflight in SQLAlchemy) adds an expert-authored substrate patch of about 250 lines on top of the public project, and one (a payments webhook pipeline) is built on a codebase owned by its author. Base commits are recent: 82 of the 100 were committed in August or September 2026, with a median date of 4 September 2026, and at least 97 lie on the upstream default branch.

\subsection{Task anatomy}\label{sec:anatomy}

Each task ships four artifacts:

\begin{itemize}
  \item \textbf{Prompt.} A ticket written by the author, prefixed with a short environment preamble giving the repository location, resource limits (about two vCPUs and a hard memory limit), the fact that the sandbox is offline, how to start any backing service the repository's own tests need, and the instruction that hidden tests will grade the work and that existing tests must not be modified (Appendix~\ref{app:prompt}).
  \item \textbf{Environment.} A Docker image containing the repository at the pinned commit, with its history vaulted so that later upstream commits cannot be read, the toolchains, and a pre-populated dependency cache. Agent shells run as an unprivileged user. Everything answer-shaped, including the reference solution, hidden grader, specification files and a pristine checkout, sits under a root-only directory that the agent cannot read.
  \item \textbf{Reference solution.} The author's own patch (\texttt{task/gold.diff}), sealed inside the image and used only for calibration and review.
  \item \textbf{Grader.} Weighted criteria, blockers and gates (\secref{sec:grading}), executed hermetically: no network, no wall-clock dependence and no randomness in any verdict. Where a criterion needs a backing service, such as Redis or a database, the grader starts its own instance inside the image on a private port.
\end{itemize}

\begin{figure}[tbp]
  \centering
  \includegraphics[width=\textwidth]{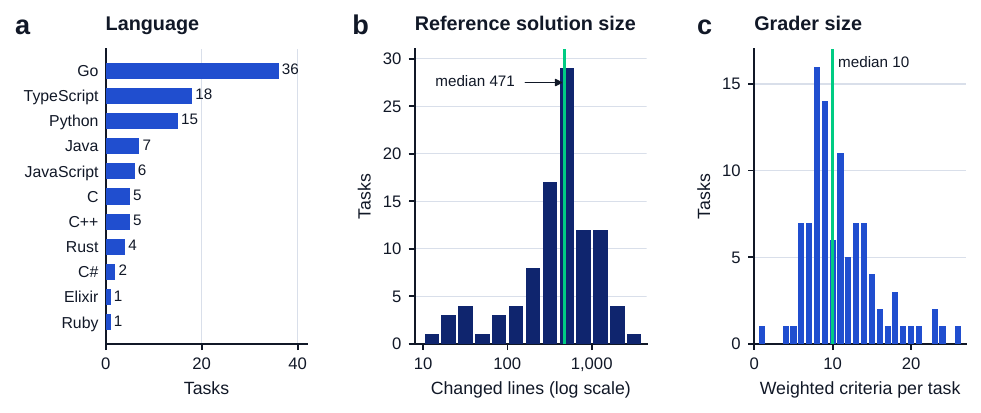}
  \caption{Composition of the 100 tasks. (a) Primary language of each task's reference change. (b) Size of the reference solution in changed lines (log scale). (c) Number of weighted criteria in each task's grader. Green lines mark medians.}
  \label{fig:corpus}
\end{figure}

Table~\ref{tab:corpus} summarizes the released corpus. Tickets are short (median 148 words) relative to the changes they request: the median reference solution changes 471 lines across 4.5 files, and agents take a median of \textbf{33 to 166 steps per run}, depending on the model (\secref{sec:effort}). The released set spans 87 open-source projects and 11 languages, and its tasks run from database engines, query builders and storage layers to schedulers, proxies, authentication servers and compilers (Figure~\ref{fig:corpus}).

\begin{table}[tbp]
  \centering\small
  \caption{Corpus statistics for the 100 released tasks. Prompt length counts the author-written ticket only, excluding the shared environment preamble. Changed lines count added plus removed lines in the reference patch.}
  \label{tab:corpus}
  \begin{tabular}{@{}lr@{}}
    \toprule
    \colhead{Property} & \colhead{Value} \\
    \midrule
    Tasks (public / held-out)                                   & 100 (24 / 76) \\
    Open-source projects / languages                            & 87 / 11 \\
    Ticket length, words: median (IQR; range)                   & 148 (69--261; 39--768) \\
    Reference solution, changed lines: median (IQR)             & 471 (270--808) \\
    Reference solution, files: median (IQR)                     & 4.5 (3--7) \\
    Agent steps per run: per-model median (range across models) & 33--166 \\
    Weighted criteria per task: median (range)                  & 10 (1--26) \\
    Tasks with gates                                            & 14 \\
    Tasks with an LLM-judged criterion (share of task weight)   & 10 (2\%--25\%) \\
    Base commits dated Aug--Sep 2026                            & 82 \\
    \bottomrule
  \end{tabular}
\end{table}

\subsection{Grading}\label{sec:grading}

Each task's grader is a list of named criteria with weights that sum to exactly 1.0; this holds for all 100 released tasks. Most criteria are executable checks of observable behaviour through the library's or service's own public interface: work enqueued and removed through the public API, documents written and read back, queries transpiled and executed. Implementation-internal checks are avoided, so that any reasonable solution can earn full credit. Graders combine three kinds of rule:

\begin{itemize}
  \item \textbf{Weighted criteria.} Partial credit is the weighted sum of criterion scores.
  \item \textbf{Blockers.} Must-pass criteria marked by the author. If a blocker fails, the run's score is capped below full credit (for example at 0.60) rather than set to zero, so that a change missing an essential property cannot score well on its other criteria.
  \item \textbf{Gates.} Pass/fail preconditions that carry no weight and cannot be traded against the criteria, for example ``the repository's own test suite is green'' and ``the repository's own tests are untouched''. If any gate fails, the run scores 0. Fourteen released tasks use gates.
\end{itemize}

Formally, for run $j$ of model $m$ on task $t$, with criterion scores $c_i \in [0,1]$, weights $w_i$ with $\sum_i w_i = 1$, blocker cap $\kappa_t \in (0,1)$ and gate set $G_t$:
\begin{equation}\label{eq:run}
  r(m,t,j) =
  \begin{cases}
    0, & \text{if any gate in } G_t \text{ fails;}\\[2pt]
    \min\Bigl(\sum_i w_i c_i,\ \kappa_t\Bigr), & \text{otherwise, if a blocker fails;}\\[2pt]
    \sum_i w_i c_i, & \text{otherwise.}
  \end{cases}
\end{equation}

\runin{Why partial credit, blockers and gates.}Pass/fail grading would compress most of the field into a narrow band. Only 1.0\% to 3.8\% of the runs of Gemini 3.8 Flash, Qwen 3.8 Max, DeepSeek V4.1 Flash and Grok 4.6 earn full credit (Table~\ref{tab:main}), so a binary score would place all four between 1\% and 4\%. With partial credit they span 32.2\% to 40.0\%, and Gemini 3.8 Flash is clearly ahead of the other three (\secref{sec:separation}). Partial credit on its own, however, would let a change that misses an essential property score well on its other criteria; blockers cap such runs below full credit. Gates encode conditions a maintainer would not trade against anything else, such as leaving the repository's own tests untouched, so a run that violates one scores zero.

\runin{LLM-judged criteria.}These criteria carry between 2\% and 25\% of their task's weight, 8.5\% on average among the ten, and 0.85\% of total benchmark weight. The remaining 90 tasks are graded entirely by deterministic checks. We kept the LLM-judged criteria to under 10\% of the total grade due to potential bias and reliability issues.

\subsection{Quality assurance and calibration}\label{sec:qa}

Quality is enforced at build time, through control patches, and through senior-engineer quality control (QC).

\begin{itemize}
  \item \textbf{Build-time calibration.} The image build grades the reference solution, the untouched checkout and an inert (comment-only) change with the real grader. The build fails unless the reference scores 1.00 with every criterion passing individually, and both the untouched checkout and the inert change score 0.00 with every criterion failing individually.
  \item \textbf{Control patches.} We have also used explicit control patches that the grader must classify correctly. These are alternative valid solutions that must pass (35 tasks), adversarial or ``cheat'' patches that must fail, for example ones that edit tests, special-case the graded inputs or disable the feature under test (30 tasks), and lazy or no-op patches that must score low (42 tasks).
  \item \textbf{Senior-engineer QC.} Every task was quality-checked by a senior engineer, who examined its prompt, environment, reference solution, grader and diagnostic frontier-model runs. Tasks that fell short were returned to their authors for revision, and only tasks that passed QC received the platform status ``verified''.
\end{itemize}

\subsection{Contamination analysis}\label{sec:contam}

The authoring rules keep each task's own change out of the public record. As an additional check, we compared every task with 499,647 task records from 93 public software engineering benchmark datasets, by matching repository commits and task identifiers and by the semantic similarity of prompts and code. An LLM judge reviewed the closest pairs. No task duplicated a public task or graded the same behaviour as one; a minority touch the same component as a public task with a different change. We also monitor the upstream repositories for changes that overlap a task's reference solution.

\subsection{Evaluation protocol}\label{sec:protocol}

\runin{Models and settings.}We evaluate seven models (Table~\ref{tab:configs}), each with extended reasoning at the platform's default high-effort setting and no other agent configuration. All evaluation jobs ran with empty agent parameters, and no system prompt was set.

\begin{table}[tbp]
  \centering\small
  \caption{Evaluated configurations. All models run in the HUD harness, which calls the provider API through the HUD inference gateway and exposes the provider's native tool interface. Reasoning effort and output-token limits are the platform defaults.}
  \label{tab:configs}
  \begin{tabularx}{\textwidth}{@{}l >{\raggedright\arraybackslash}p{2.9cm} L l r@{}}
    \toprule
    \colhead{Model} & \colhead{Agent / API} & \colhead{Native tools exposed} & \colhead{Reasoning} & \colhead{Max output} \\
    \midrule
    Claude Opus 5.5     & Anthropic Messages     & \tool{bash}, \tool{str\_replace\_based\_edit\_tool} & effort max    & 128,000 \\
    Claude Fable 5.1    & Anthropic Messages     & \tool{bash}, \tool{str\_replace\_based\_edit\_tool} & effort max    & 128,000 \\
    GPT-6 Astra         & OpenAI Responses       & \tool{shell} (patch-based edits) & effort max    & 64,000 \\
    Gemini 3.8 Flash    & Gemini generate\-Content & \tool{run\_shell\_command}, \tool{read\_file}, \tool{replace}, \tool{write\_file}, \tool{grep\_search}, \tool{glob}, \tool{list\_directory} & thinking high & 65,536 \\
    Qwen 3.8 Max        & OpenAI-compatible Chat & \tool{bash}, \tool{read}, \tool{edit}, \tool{write}, \tool{grep}, \tool{glob} & effort xhigh  & 131,072 \\
    DeepSeek V4.1 Flash & OpenAI-compatible Chat & \tool{bash}, \tool{read}, \tool{edit}, \tool{write}, \tool{grep}, \tool{glob} & effort xhigh  & 131,072 \\
    Grok 4.6            & OpenAI-compatible Chat & \tool{bash}, \tool{read}, \tool{edit}, \tool{write}, \tool{grep}, \tool{glob} & effort xhigh  & \makecell[tr]{provider\\default} \\
    \bottomrule
  \end{tabularx}
\end{table}

\runin{Harness.}Every run uses the HUD hosted runner and the HUD SDK (\texttt{hud-python}, v0.6 series; latest release v0.6.21). We chose HUD because it runs every provider's models through one harness against the same container and grader, enforces the build-time calibration described in \secref{sec:qa}, and records a full trace of every run for audit. The SDK's agent loop calls the model's API directly; no vendor product such as Claude Code, Codex CLI or Gemini CLI is involved. The loop exposes the tool schema each provider's models are trained on (Table~\ref{tab:configs}), and every tool call executes in the same task container. All models therefore share the task, prompt, container, resources, step budget and grader, and differ only in tool interface and agent loop, which the SDK fixes per provider. This differs from DeepSWE's choice of a single bash tool for every model \citep{huang2026deepswe}; we discuss the trade-off in \secref{sec:limitations}.

\runin{Budget and sampling.}Each run is limited to 1,000 agent steps. Each model attempts each task five times. Only the five most recent valid, completed runs per model and task count toward the score. Runs are excluded and rerun if they were invalidated on the platform, failed for infrastructure reasons, were smoke checks, or hit the step limit or output-token limit. Limit-hit exclusions were concentrated in two models: 27 Claude Opus 5.5 runs and 55 Gemini 3.8 Flash runs, plus 2 Grok 4.6 runs. Gemini 3.8 Flash has no valid run on one released task and is scored over 99 tasks. The final dataset contains 3,495 scored runs, collected between 1 and 10 October 2026.

\subsection{Metrics and uncertainty}\label{sec:metrics}

A run's score $r(m,t,j) \in [0,1]$ is defined in Eq.~\eqref{eq:run} (\secref{sec:grading}). A model's task score is the mean over its $n_{m,t}$ scored runs on that task ($n_{m,t} = 5$ for every model and task), and its benchmark score is the macro-average over the $N_m$ tasks it covers ($N_m = 100$, or 99 for Gemini 3.8 Flash), so that every task counts equally:
\begin{equation}\label{eq:score}
  s(m,t) = \frac{1}{n_{m,t}} \sum_{j=1}^{n_{m,t}} r(m,t,j), \qquad
  S(m) = \frac{1}{N_m} \sum_{t} s(m,t).
\end{equation}

We report three kinds of uncertainty metrics, because they answer different questions:

\begin{itemize}
  \item \textbf{Task-sampling intervals} (headline): $S \pm 1.96 \cdot \mathrm{SE}$, where SE is the standard error of the task scores. This asks how much the score would move on a different but equally reasonable set of tasks, and it is the interval shown on the public leaderboard.
  \item \textbf{Paired tests} for differences between models. Tasks are resampled 2,000 times to give a bootstrap interval for the mean per-task difference, and a sign-flip permutation test (20,000 permutations, Holm-adjusted across comparisons) gives $p$-values against the top model. Because both models are scored on the same tasks, paired tests are the appropriate basis for a ranking claim.
  \item \textbf{Run-to-run intervals}, for comparability with DeepSWE and Terminal-Bench \citep{huang2026deepswe,merrill2026terminalbench}: the benchmark is scored separately on each of the five run slots, and the interval is $1.96 \cdot \operatorname{std}(S_1, \dots, S_5)/\sqrt{5}$. This captures only sampling noise in the agent's behaviour on a fixed task set.
\end{itemize}

We also report reliability measures: the mean of each model's best run out of five, the share of runs that earn full credit, the number of tasks with at least one full-credit run out of five (an analogue of pass@5), the number with five out of five, and the mean within-task standard deviation of run scores. Steps are agent steps, meaning model turns.

\section{Results}\label{sec:results}

\subsection{Leaderboard}\label{sec:leaderboard}

\begin{figure}[tbp]
  \centering
  \includegraphics[width=\textwidth]{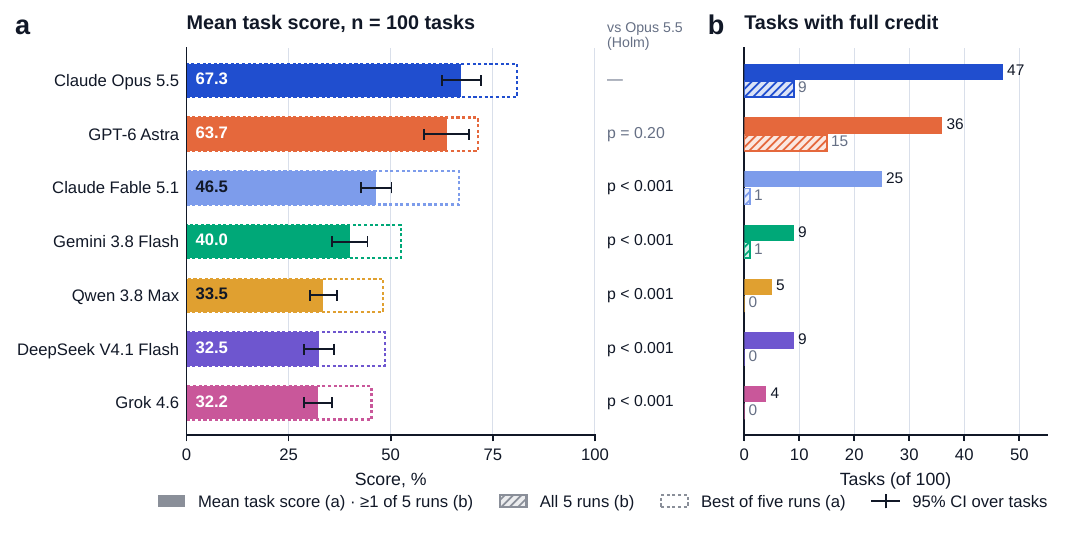}
  \caption{EC-SWE leaderboard. (a) Mean task score on the 100 tasks (solid bars), with 95\% intervals over tasks. Dotted outlines show the mean of each model's best run out of five. The right-hand column gives Holm-adjusted $p$-values from paired sign-flip permutation tests against Claude Opus 5.5. (b) Number of tasks on which a model earned full credit on at least one of its five runs (solid) and on all five runs (hatched). Every model was run five times per task with a 1,000-step limit.}
  \label{fig:leaderboard}
\end{figure}

Table~\ref{tab:main} and Figure~\ref{fig:leaderboard} report each model's mean task score on the 100 released tasks. Claude Opus 5.5 scores 67.3\% and GPT-6 Astra 63.7\%, followed by Claude Fable 5.1 (46.5\%) and Gemini 3.8 Flash (40.0\%, 99 tasks). Qwen 3.8 Max, DeepSeek V4.1 Flash and Grok 4.6 form a lower group between 32.2\% and 33.5\%. The mean across all seven models is 45.1\%. If only each model's best run out of five counted, the scores would rise to 80.9\% for Claude Opus 5.5, 71.4\% for GPT-6 Astra and 66.7\% for Claude Fable 5.1 (dotted outlines in Figure~\ref{fig:leaderboard}a); the gain is largest for the least consistent models (\secref{sec:reliability}). These are the scores published with the benchmark release.

\begin{table}[tbp]
  \centering\small
  \begin{threeparttable}
  \caption{Main results on the 100 released tasks, 3,495 scored runs. Score is the macro-average of task scores, with a 95\% interval over tasks; the run-to-run interval covers agent sampling noise only (\secref{sec:metrics}). ``Full-credit runs'' is the share of runs scoring 1.0. ``$\geq$1 of 5'' and ``5 of 5'' count tasks with at least one, or all five, full-credit runs. ``Within-task SD'' is the mean standard deviation of run scores within a task, in points.}
  \label{tab:main}
  \begin{tabular}{@{}lccccc@{}}
    \toprule
    \colhead{Model} & \colhead{Score (95\% CI)} & \colhead{Run-to-run CI} & \colhead{Full-credit runs} & \makecell{\colhead{Tasks $\geq$1 of 5}\\\colhead{/ 5 of 5}} & \colhead{Within-task SD} \\
    \midrule
    Claude Opus 5.5                & $\mathbf{67.3 \pm 4.7}$ & $\pm 1.9$ & 26.6\% & 47 / 9  & 13.6 \\
    GPT-6 Astra                    & $63.7 \pm 5.5$ & $\pm 1.5$ & 23.4\% & 36 / 15 & 7.3 \\
    Claude Fable 5.1               & $46.5 \pm 3.7$ & $\pm 0.4$ & 8.6\%  & 25 / 1  & 16.9 \\
    Gemini 3.8 Flash\tnote{a}      & $40.0 \pm 4.3$ & $\pm 1.5$ & 3.8\%  & 9 / 1   & 11.0 \\
    Qwen 3.8 Max                   & $33.5 \pm 3.3$ & $\pm 1.4$ & 1.0\%  & 5 / 0   & 10.9 \\
    DeepSeek V4.1 Flash            & $32.5 \pm 3.7$ & $\pm 1.4$ & 3.0\%  & 9 / 0   & 13.0 \\
    Grok 4.6                       & $32.2 \pm 3.5$ & $\pm 1.2$ & 1.0\%  & 4 / 0   & 10.1 \\
    \bottomrule
  \end{tabular}
  \begin{tablenotes}\footnotesize
    \item[a] Scored over 99 tasks; Gemini 3.8 Flash has no valid run on one released task.
  \end{tablenotes}
  \end{threeparttable}
\end{table}

The task-sampling intervals ($\pm$3.3 to $\pm$5.5 points) are two to ten times wider than the run-to-run intervals ($\pm$0.4 to $\pm$1.9). On a fixed set of 100 tasks, rerunning the agents barely moves the scores; most of the uncertainty comes from which tasks are in the benchmark. Leaderboards that report only run-to-run intervals therefore understate how much a ranking depends on the particular task sample \citep{madaan2024variance}.

\subsection{Statistical separation}\label{sec:separation}

Because all models are scored on the same tasks, paired comparisons are much sharper than overlapping marginal intervals suggest (Table~\ref{tab:paired}). They resolve four tiers. Against the top model, a paired permutation test gives $p = 0.20$ for GPT-6 Astra and $p < 0.001$ for every other model (Holm-adjusted; Figure~\ref{fig:leaderboard}a), i.e., Claude Opus 5.5 outperforms those models in a statistically significant manner. The top two models are not separable: Claude Opus 5.5 leads GPT-6 Astra by 3.6 points, but the paired 95\% interval $[-1.4, +9.7]$ includes zero, and each model wins a large share of tasks (53 against 41, with 6 ties). Both are clearly ahead of Claude Fable 5.1, which Claude Opus 5.5 outscores on 82 of 100 tasks. Fable is in turn ahead of Gemini 3.8 Flash ($+6.5$ $[+1.6, +10.5]$), and Gemini is ahead of the lower three ($+6.5$ over Qwen 3.8 Max $[+3.0, +11.0]$). Qwen 3.8 Max, DeepSeek V4.1 Flash and Grok 4.6 cannot be distinguished from one another; every pairwise interval includes zero.

\begin{table}[tbp]
  \centering\small
  \caption{Paired differences between models on shared tasks. $\Delta$ is the difference in mean task score in points (first model minus second), with a 95\% paired bootstrap interval over tasks (2,000 resamples). W\,/\,T\,/\,L counts tasks where the first model scores higher, ties, or scores lower.}
  \label{tab:paired}
  \begin{tabular}{@{}lccc@{}}
    \toprule
    \colhead{Comparison} & \colhead{$\boldsymbol{\Delta}$ (points)} & \colhead{95\% paired CI} & \colhead{W / T / L} \\
    \midrule
    Claude Opus 5.5 $-$ GPT-6 Astra         & $+3.6$  & $[-1.4, +9.7]$   & 53 / 6 / 41 \\
    Claude Opus 5.5 $-$ Claude Fable 5.1    & $+20.8$ & $[+16.9, +25.3]$ & 82 / 5 / 13 \\
    GPT-6 Astra $-$ Claude Fable 5.1        & $+17.2$ & $[+11.8, +22.1]$ & 70 / 2 / 28 \\
    Claude Fable 5.1 $-$ Gemini 3.8 Flash   & $+6.5$  & $[+1.6, +10.5]$  & 63 / 3 / 33 \\
    Gemini 3.8 Flash $-$ Qwen 3.8 Max       & $+6.5$  & $[+3.0, +11.0]$  & 50 / 6 / 43 \\
    Qwen 3.8 Max $-$ DeepSeek V4.1 Flash    & $+1.0$  & $[-2.2, +3.5]$   & 64 / 4 / 32 \\
    DeepSeek V4.1 Flash $-$ Grok 4.6        & $+0.3$  & $[-3.4, +3.8]$   & 45 / 6 / 49 \\
    Qwen 3.8 Max $-$ Grok 4.6               & $+1.3$  & $[-2.5, +4.7]$   & 53 / 5 / 42 \\
    \bottomrule
  \end{tabular}
\end{table}

\subsection{Task-level difficulty and headroom}\label{sec:difficulty}

\begin{figure}[tbp]
  \centering
  \includegraphics[width=\textwidth]{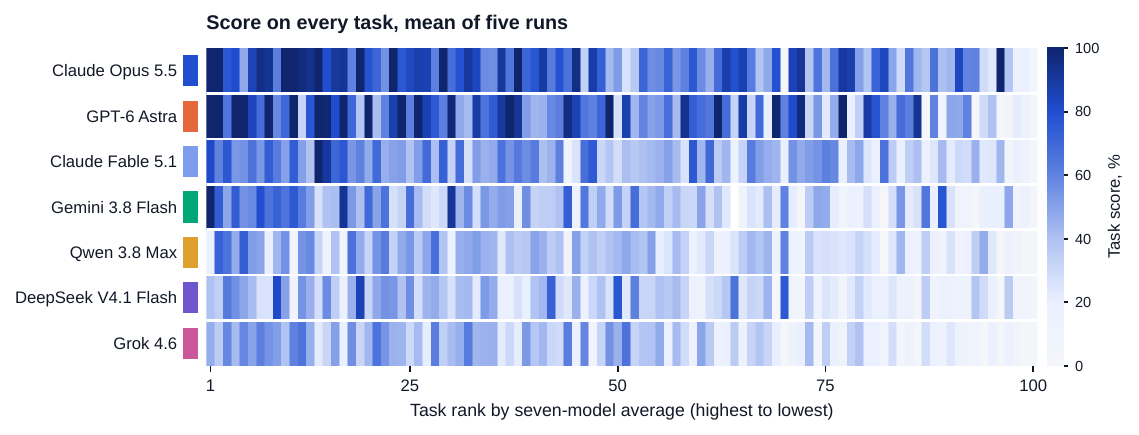}
  \caption{Task-by-model score matrix for the 100 tasks, ordered left to right by seven-model average. Each cell is the mean of five runs. The empty cell in the Gemini 3.8 Flash row is the one task without a valid Gemini run. Colour swatches identify models consistently across all figures.}
  \label{fig:heatmap}
\end{figure}

Figure~\ref{fig:heatmap} shows every model's score on every task. The benchmark has substantial headroom. No task is fully solved by all seven models. On 80 tasks, no model earns full credit on all five runs, and the two leading models both score 100\% on only four tasks while both falling below 60\% on 20. Task averages across the seven models range from 2.5\% to 69.5\% (median 46.2\%), and on 9 tasks even the best model averages below 50\%. Of the 699 model--task pairs, only 26 earn full credit on every run and 14 earn zero on every run.

Difficulty is far from uniform across models, and the two leaders are complementary rather than ordered. Their task scores correlate only moderately (Pearson $r = 0.42$), and on 14 tasks they differ by 40 points or more, 8 in Claude Opus 5.5's favour and 6 in GPT-6 Astra's.

A few tasks are hard for every model. In an OAuth2 Proxy task on session and CSRF cookie accounting, no model averages above 21.8\%. In a Harbor quota-ownership task, the best model averages 19.7\%. On the hardest task, a SystemVerilog export for the CirC compiler, all seven models land between 2.3\% and 2.7\%.

\subsection{Run-level outcomes and reliability}\label{sec:reliability}

\begin{figure}[tbp]
  \centering
  \includegraphics[width=\textwidth]{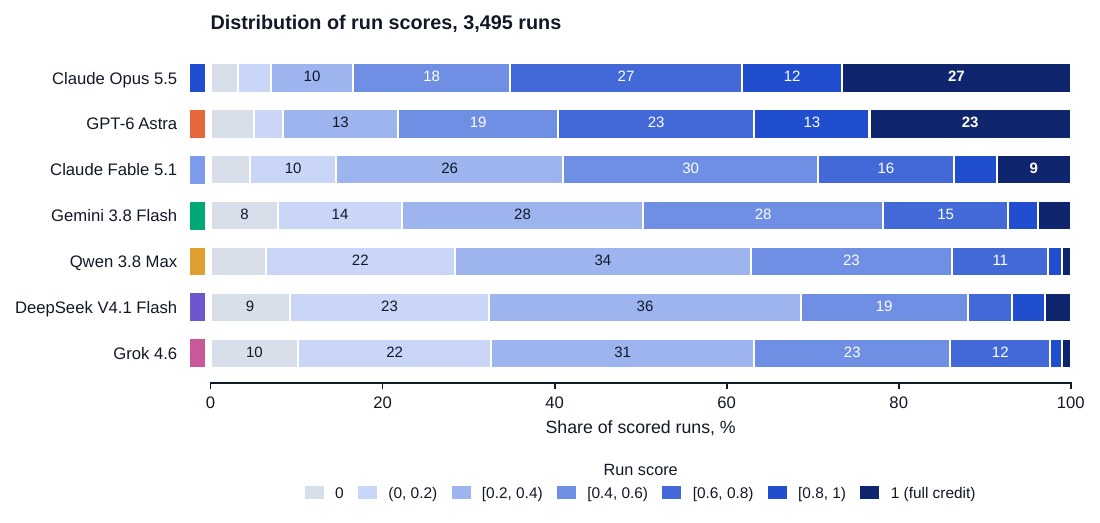}
  \caption{Distribution of run scores across all scored runs for each model. Segments are labelled with their percentage where space allows.}
  \label{fig:runs}
\end{figure}

Partial credit shows how close agents get when they fall short (Figure~\ref{fig:runs}). Even for the two strongest models, most runs earn at least 0.4 (83\% for Claude Opus 5.5, 78\% for GPT-6 Astra), but only 26.6\% of Claude Opus 5.5 runs and 23.4\% of GPT-6 Astra runs earn full credit, against at most 8.6\% for every other model. The lower group rarely earns full credit at all (1.0\% of Qwen 3.8 Max and Grok 4.6 runs); its typical run collects between a fifth and a half of a task's weight.

The leaders differ in consistency as well as in level. GPT-6 Astra's runs vary less within a task (mean within-task SD 7.3 points against 13.6 for Claude Opus 5.5). It reaches full credit on all five runs for 15 tasks against Claude Opus 5.5's 9, while Claude Opus 5.5 reaches full credit at least once on more tasks (47 against 36). A user who can sample several attempts and verify them would favour Claude Opus 5.5's breadth; a user who needs a single attempt to be right would value GPT-6 Astra's reliability. Claude Fable 5.1 is the least consistent model (within-task SD 16.9) even though its run-to-run benchmark score is the most stable ($\pm$0.4). Its variability averages out across tasks.

\subsection{Where credit is lost: blockers and gates}\label{sec:gates}

Blockers and gates encode the conditions under which a maintainer would reject a change regardless of its other merits. Across all 3,495 scored runs, 274 runs (7.8\%) were capped by a failed blocker and 80 runs (2.3\%) were zeroed by a failed gate (Table~\ref{tab:gates}). These rules cost the strongest models the most, because their uncapped criteria scores are highest. Without blocker caps and gates, Claude Opus 5.5 would gain 3.7 points and GPT-6 Astra 2.9, while the lower group would gain between 1.1 and 2.9. The ranking would not change.

\begin{table}[tbp]
  \centering\small
  \caption{Credit lost to blockers and gates. ``Capped'' counts runs whose score was reduced by a failed blocker; ``zeroed'' counts runs set to zero by a failed gate. ``Mean loss'' is the average difference between the uncapped criteria score and the final score for those runs. ``Points lost'' is the total effect on the benchmark score.}
  \label{tab:gates}
  \begin{tabular}{@{}lccccc@{}}
    \toprule
    \colhead{Model} & \colhead{Capped runs} & \makecell{\colhead{Mean loss}\\\colhead{(capped)}} & \colhead{Zeroed runs} & \makecell{\colhead{Mean loss}\\\colhead{(zeroed)}} & \colhead{Points lost} \\
    \midrule
    Claude Opus 5.5     & 62 (12.4\%) & 0.22 & 11 (2.2\%) & 0.43 & 3.7 \\
    GPT-6 Astra         & 42 (8.4\%)  & 0.26 & 12 (2.4\%) & 0.30 & 2.9 \\
    Claude Fable 5.1    & 37 (7.4\%)  & 0.27 & 12 (2.4\%) & 0.18 & 2.4 \\
    Gemini 3.8 Flash    & 31 (6.3\%)  & 0.25 & 8 (1.6\%)  & 0.00 & 1.6 \\
    Qwen 3.8 Max        & 29 (5.8\%)  & 0.21 & 10 (2.0\%) & 0.00 & 1.2 \\
    DeepSeek V4.1 Flash & 24 (4.8\%)  & 0.23 & 9 (1.8\%)  & 0.03 & 1.1 \\
    Grok 4.6            & 49 (9.8\%)  & 0.22 & 18 (3.6\%) & 0.21 & 2.9 \\
    \bottomrule
  \end{tabular}
\end{table}

The gates catch behaviour that test-only grading would reward. The clearest case is an Asynq task on how task cancellation interacts with handler outcomes. Claude Opus 5.5 and Grok 4.6 score zero on every run, while DeepSeek V4.1 Flash averages 75.6\%.

\subsection{Effort}\label{sec:effort}

\begin{figure}[tbp]
  \centering
  \includegraphics[width=0.74\textwidth]{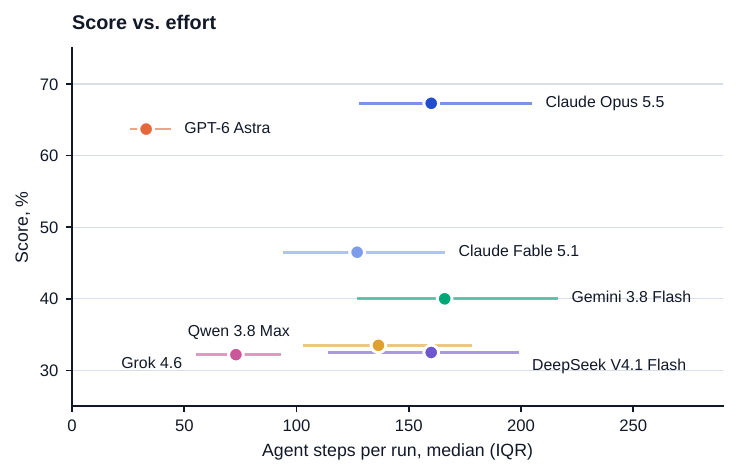}
  \caption{Score against median agent steps per run; whiskers show the interquartile range of steps across each model's runs.}
  \label{fig:effort}
\end{figure}

Agents take very different routes to their scores (Table~\ref{tab:effort}, Figure~\ref{fig:effort}). The median number of agent steps per run \textbf{varies five-fold}, from 33 for GPT-6 Astra to 166 for Gemini 3.8 Flash, and five of the seven models take a median of 127 steps or more (160 for Claude Opus 5.5). Step count is a poor predictor of success: GPT-6 Astra reaches 95\% of Claude Opus 5.5's score with about a fifth of its steps, and Gemini 3.8 Flash takes the most steps of any model while scoring fourth. Within each model, the correlation between a run's step count and its score is weak ($r$ between $-0.15$ and $+0.17$). Wall-clock time follows yet another pattern: Qwen 3.8 Max has the longest median container lifetime (100 minutes) despite a mid-range step count, and GPT-6 Astra the shortest (36 minutes).

\begin{table}[tbp]
  \centering\small
  \caption{Effort per scored run. Steps are agent steps (model turns). Container time is the wall-clock lifetime of the task container, including grading.}
  \label{tab:effort}
  \begin{tabular}{@{}lcccc@{}}
    \toprule
    \colhead{Model} & \colhead{Score} & \colhead{Median steps (IQR)} & \colhead{Mean steps} & \colhead{Median container time} \\
    \midrule
    Claude Opus 5.5     & 67.3 & 160 (128--205) & 171 & 80 min \\
    GPT-6 Astra         & 63.7 & 33 (26--44)    & 37  & 36 min \\
    Claude Fable 5.1    & 46.5 & 127 (94--166)  & 134 & 56 min \\
    Gemini 3.8 Flash    & 40.0 & 166 (127--217) & 182 & 40 min \\
    Qwen 3.8 Max        & 33.5 & 137 (103--178) & 147 & 100 min \\
    DeepSeek V4.1 Flash & 32.5 & 160 (114--199) & 162 & 51 min \\
    Grok 4.6            & 32.2 & 73 (55--93)    & 80  & 69 min \\
    \bottomrule
  \end{tabular}
\end{table}

\section{Discussion and Conclusion}\label{sec:discussion}

\subsection{What the results show}\label{sec:findings}

\runin{Expert-authored, rubric-graded tasks separate frontier agents.}On EC-SWE, seven current models span 35 points and fall into four statistically distinct tiers. The same models sit much closer together on DeepSWE v1.1, where four of them (Claude Opus 5.5, GPT-6 Astra, DeepSeek V4.1 Flash and Gemini 3.8 Flash) fall within 0.9 points of one another \citep{deepswe2026leaderboard}; on EC-SWE those four span 34.8 points. The two benchmarks agree on the top two models but diverge below them, most clearly for DeepSeek V4.1 Flash, which ranks third on DeepSWE and sixth here. The benchmarks differ in metric, harness and task population, so this comparison is indicative rather than controlled, but it suggests that EC-SWE resolves differences that existing leaderboards do not. The likely sources are the task design itself (changes that combine two kinds of work in a codebase the agent has never seen), partial credit, which distinguishes ``most of the change'' from ``little of it'', and maintainer-style rules that refuse to reward changes a reviewer would reject.

\runin{The two leading models are complementary.}Claude Opus 5.5 and GPT-6 Astra are statistically tied on average but often disagree on individual tasks: their task scores correlate at only $r = 0.42$, and on 14 tasks they differ by 40 points or more. They also trade breadth for consistency, and they differ greatly in effort. A single leaderboard number hides all of this. We therefore recommend reading EC-SWE alongside its task-level, reliability and effort views.

\runin{Effort is not capability.}Step count does not track score: the model that takes the fewest steps is the second-best overall, and the one that takes the most scores fourth. For deployments where latency matters, effort should be reported next to accuracy, as recent benchmarks also argue \citep{huang2026deepswe}.

\runin{Maintainer rules matter.}About one run in ten loses credit to a blocker or gate, and the strongest models lose the most. Some of these losses are exactly the behaviour test-only grading rewards: editing the repository's own tests to make a suite pass, or shipping a change that breaks a property the ticket said must be kept.

\subsection{Limitations}\label{sec:limitations}

\begin{itemize}
  \item \textbf{Harness.} Each model gets its provider's native tool interface through the HUD SDK. This is closer to how the models are trained and deployed than a shared bash tool, but it means the agent loop and tools are not identical across providers. Part of a gap between models could therefore reflect tool design rather than model capability. A controlled comparison against a single shared harness is future work.
  \item \textbf{Reasoning settings.} Each model runs at its platform default high-effort setting (max, xhigh or high, depending on the provider). We did not sweep reasoning effort, so other settings could change both scores and effort.
  \item \textbf{Sample size.} One hundred tasks give task-sampling intervals of $\pm$3.3 to $\pm$5.5 points. Differences smaller than about 5 points between adjacent models should not be over-interpreted.
  \item \textbf{Contamination.} Task changes are unpublished and their similarity to 93 public benchmark datasets was checked, but the repositories themselves are public and their code is likely in training data. EC-SWE tests problem solving in familiar-looking code, not in code no model has ever seen. Released public tasks will eventually enter training corpora; the 76 held-out tasks mitigate this.
\end{itemize}

\subsection{Future work}\label{sec:future}

We plan to (i) audit graders with an independent judge on a sample of agent runs and publish disagreement rates; (ii) rerun all models under a shared single-tool harness to separate tool effects from model effects; (iii) sweep reasoning effort to map score--effort trade-offs; (iv) grow the corpus through further authoring waves and periodically retire public tasks; and (v) extend the same expert-authored, rubric-graded methodology to other professional domains, starting with management consulting.

\subsection{Conclusion}\label{sec:conclusion}

EC-SWE asks frontier coding agents to make changes that experienced engineers would ask for, in codebases those engineers know, and grades the results the way a maintainer would: with partial credit for what works, caps for missing essentials, and zero for changes that should never be merged. On 100 such tasks, seven frontier models span 35 points, the top two are statistically tied but complementary, and no task is solved by every model. We release 24 tasks with their environments, graders and results, and hold out 76, so that the community can inspect the method and reproduce the measurement.

\subsection*{Acknowledgements}

We thank the ExpertCrowd talent, the engineers of Toptal's ExpertCrowd network who designed, built and verified every task in this benchmark; EC-SWE exists because of their expertise and care. We also thank the HUD team for their contribution to this work, including the evaluation platform, environment templates and tooling on which every task was built and every run was executed. Finally, we thank the maintainers of the open-source projects whose code the tasks build on.

\nocite{*}
\bibliographystyle{plainnat}
\bibliography{refs}

\clearpage
\appendix
\titleformat{\section}{\sffamily\Large\bfseries\color{ecnavy}}{Appendix \thesection}{0.75em}{}

\section{Prompt template}\label{app:prompt}

Every task prompt consists of an environment preamble followed by the author's ticket. The preamble below is reproduced from one task, with task-specific details elided ([\dots]). Authors adapt the environment sentences to their repository; the opening and closing instructions follow a common template.

\begin{tcolorbox}[enhanced,colback=eclight,colframe=eccool,boxrule=0.5pt,arc=2pt,
                  left=8pt,right=8pt,top=6pt,bottom=6pt,before skip=10pt]
\begin{Verbatim}[fontsize=\small,breaklines=true]
You are working in a coding repository located at /workspace/repo. Please make
your changes directly in /workspace/repo, not in a copy or a separate worktree.
Environment: this container has about 2 vCPUs and a hard memory limit, and a
fully parallel build or test run can exhaust it. Build and test package by
package, e.g. [...]. Treat the sandbox as offline: the toolchain runs against a
pre-populated module cache, so builds and tests work but nothing can be fetched.
[...how to start any backing service the repository's own tests need...]

Use the tools provided to complete the following task. Hidden tests grade your
work when you finish; do not modify existing tests.

# Ticket
<author-written ticket>
\end{Verbatim}
\end{tcolorbox}

\end{document}